\documentclass[runningheads]{llncs}
\usepackage{graphicx}

\usepackage{tikz}
\usepackage{comment}
\usepackage{amsmath,amssymb} 
\usepackage{color}
\usepackage[colorinlistoftodos,prependcaption,textsize=tiny]{todonotes}

\begin{document}
\pagestyle{headings}
\mainmatter
\def\ECCVSubNumber{16} 

\title{Score-level Multi Cue Fusion for Sign Language Recognition} 

\titlerunning{Score-level Multi Cue Fusion}
%
\author{\c{C}a\u{g}r{\i} G{\"o}k\c{c}e\inst{1} \and
O\u{g}ulcan {\"O}zdemir\inst{1} \and
Ahmet Alp K{\i}nd{\i}ro\u{g}lu\inst{1} \and
Lale Akarun\inst{1}
}
\authorrunning{C. G{\"o}k\c{c}e et al.}
%
\institute{Bo\u{g}azi\c{c}i University, Computer Engineering Department, Istanbul, Turkey
\email{\{cagri.gokce, ogulcan.ozdemir, alp.kindiroglu, akarun\}@boun.edu.tr}}

\maketitle

\begin{abstract}
Sign Languages are expressed through hand and upper body gestures as well as facial expressions. Therefore, Sign Language Recognition (SLR) needs to focus on all such cues. Previous work uses hand-crafted mechanisms or network aggregation to extract the different cue features, to increase SLR performance. This is slow and involves complicated architectures. We propose a more straightforward approach that focuses on training separate cue models specializing on the dominant hand, hands, face, and upper body regions. We compare the performance of 3D Convolutional Neural Network (CNN) models specializing in these regions, combine them through score-level fusion, and use the weighted alternative. Our experimental results have shown the effectiveness of mixed convolutional models. Their fusion yields up to $19\%$ accuracy improvement over the baseline using the full upper body. Furthermore, we include a discussion for fusion settings, which can help future work on Sign Language Translation (SLT).

\end{abstract}

\section{Introduction}

Sign Language is the means of communication of the Deaf, and each Deaf culture has its own sign language. Sign languages differ from the spoken language of the culture. Communication between the Deaf and the hearing relies mostly on the Deaf individual learning the spoken language and using lipreading and written text to communicate: A huge and unfair burden on the Deaf. The reverse, teaching the general population at least some sign language may be more feasible, and there are available educational courses for such aim. However, gaining expertise in sign language is difficult, and the communication problem is still unsolved. Automatic interpretation of sign languages is a necessary step for not only enabling the human-computer interaction but also facilitating the communication between the Deaf and the hearing individuals.

Automatic Sign Language Recognition (ASLR) refers to a broad field with different tasks, such as recognizing isolated sign glosses and continuous sign sentences. The objective of the ASLR system is to infer the meaning of the sign glosses or sentences and translate it to the spoken language. Recently, there has been an increased progress in these efforts: Sign Language Translation (SLT) has become an active research problem for creating interactive sign language interfaces for the deaf \cite{camgoz2020multi,camgoz2018slt,camgoz2020transformers,orbay2020neural}. A number of recent papers on the topic made use of neural network generated features. However, while the quality and representative power of these features in SLT are essential, and it is difficult to evaluate the representative potential of the elements in a pipeline setting where the overall system error is cumulative. For this reason, in this study, we aim to evaluate 3D Residual CNN Based Sign Language embeddings in terms of explanatory power in an Automatic Sign Language Recognition (ASLR) setting where temporal mix-up between signs and co-articulation is minimal. For the general case of Isolated SLR, the system aims to process a sign gloss and assign it to a single sign gloss label. In a limited context of supervised learning set-up, labels are glosses, which are transcription symbols assigned by sign language experts. There may be a single signer or multiple signers in communication; however, the ASLR system should be signer independent.

To convey the meaning of a performed sign gloss, Sign Languages use multiple channels, which are manifested as visual cues. We can classify these visual cues into two categories; (1) cues that are denoted as manual cues including hand shape and movement, and (2) cues that are non-manual features including facial expressions and upper body pose focusing on details without definitive large displacements. 

Solving the problem of Isolated SLR requires specialized methods, which can be grouped into two categories. The first category is using handcrafted features, focusing on video trajectories and flow maps ~\cite{zhang2016csl,ozdemir2020bosphorussign22k,wang2013action}. The second set of methods includes machine learning algorithms and neural networks to improve classification performance ~\cite{li2020wlasl,joze2019msasl,ozdemir2020bosphorussign22k}. 3D CNN models have proven successful in various video tasks \cite{tran2015learning,tran2018resnet3d}. Li et al.~\cite{li2020wlasl} adopted the same architecture in SLR and reported improved performance. However, {\"O}zdemir et al.~\cite{ozdemir2020bosphorussign22k} provided the comparison of 3D CNN models and handcrafted methods but have found that 3D CNNs are inferior to the state-of-the-art handcrafted IDT approach.

The aim of this work is to investigate why 3D CNN models may fail to show similar success in sign language recognition and to observe what modifications improve their performance. We hypothesize that the performance drop occurs because of the common practice of scaling images into smaller size and sampling frames~\cite{tran2015learning,tran2018resnet3d}, due to computational requirements and difficulty of training bigger neural networks. One solution is handling the negative effect of the sampling by increasing the model complexity as in ~\cite{feichtenhofer2019slowfast,zhou2020spatial,kindiroglu2019taf}, yet this increases computational requirements. Instead, we firstly apply attentive data selection at the pre-processing phase by determining cues in SLR data. Secondly, we divide the problem into multiple cues and train different expert classifiers on each kind of dense feature. Thirdly, we refine the expert cue network knowledge into one result, by applying score-level fusion.

The paper organization is as follows. Sections \ref{sec2:related_work} reviews related work,  Section \ref{sec3:methodology} explains the presented method, Section 4 presents the experimental results,  Section \ref{sec5:discussion} contains the analysis of experiments and  Section \ref{sec6:conclusion} presents the conclusions.

\section{Related Work} \label{sec2:related_work}

Sign Language Recognition (SLR) aims to infer meaning from a performed sign. In the sign classification task, an isolated sign gloss is assigned to a class label. A sign gloss, the written language counterpart of the performed sign, can be used as a mid-level or final stage label for sign language recognition.

SLR is closely connected with video recognition or human action recognition methods, and similar architectures have been used for both. Two popular approaches to sign language representation uses handcrafted features and deep neural network based methods.

Prior to the performance leap achieved by neural networks, hand-crafted features were the best performing approach for representing human actions in a sequential video setting. For a two-frame dynamic flow map estimation, optical flow is used to generate feature-level information. These features perform better representation than RGB image sequences where the motion information is more indicative than appearance \cite{carreira2017inflated}. There exist numerous handcrafted feature extraction methods and their application to image sequences such as STIP \cite{li2017survey} and spatio-temporal local binary patterns \cite{wang2015efficient}. State of the art performances with constructed features in action recognition and isolated sign language recognition were obtained using Improved Dense Trajectories \cite{wang2013action,ozdemir2020bosphorussign22k}, which is an outlier independent trajectory-based motion specialized feature extractor.

Neural Network based methods focus on convolutional architectures for the classification task. Simonyan et al.~\cite{simonyan2014two} use a branched CNN architecture that splits the information into spatial and temporal streams, and fuses them to perform video classification. Tran et al.~\cite{tran2015learning} use 3D convolutional kernels to build a 3D CNN variant to process video data in an end-to-end fashion. 

One prerequisite for using deep neural networks is the presence of large datasets with ground truth annotations. Recently, big-scale isolated sign language recognition datasets have become publicly available. Isolated SL datasets contain videos of a user performing a single gloss, usually a single word or a phrase. MS-ASL~\cite{joze2019msasl} is an American Isolated SL dataset including 200 native performers performing more than a thousand word categories. WL-ASL~\cite{li2020wlasl} is a bigger dataset with two thousand word categories performed by one hundred people. For other languages, Chinese~\cite{zhang2016csl} and Turkish~\cite{ozdemir2020bosphorussign22k} are among available datasets. Popular human activity recognition datasets ~\cite{soomro2012ucf101,kuehne2011hmdb,karpathy2014large,kay2017kinetics} are also used as extra data and for finetuning in Isolated SLR. Continuous SL datasets are acquired in a less controlled setting, where a user can perform longer sign sequences ~\cite{hanke2010dgs,forster2014extensions}.

SLR methods often use video pre-processing to reduce network bias and variance, and to increase network performance. Random cropping is one of the popular spatial augmentation techniques when training CNNs. Since CNN variants have small input spatial resolution, e.g., $224\times224$ for the popular ResNet50 network~\cite{he2016resnet}, such methods increase the transitive invariance of the models by processing different parts of the image in higher resolution compared to directly downsampling the whole image frame. 

Temporal pre-processing techniques operate on the temporal dimension of the video data. The aim is to locate the dense temporal regions which have an increased likelihood of the action flow. In recent work, different approaches are applied for the temporal activity localization, e.g., exploiting both short term and long term samples~\cite{varol2017long}, combining high and low-frequency learners~\cite{feichtenhofer2019slowfast}, and detecting active window boundaries for the long sequences~\cite{lin2019bmn}. Our work differs by applying cue selection before the training phase and combining the classifiers in the feature construction stage.

Combining both pre-processing techniques allows an opportunity to exploit covariance between these spatio-temporal features. Spatio-temporal pre-processing can possibly improve the signal to noise ratio of the processed data when the region of interest is selected from dense regions. This process is shown to be beneficial on other video recognition tasks, e.g., when extracted through handcrafted methods such as optical flow~\cite{simonyan2014two}, or directly through 3D CNNs~\cite{tran2018resnet3d}. In SLR, due to the nature of the task, SL videos consist of the sparse hand and upper body movements as well as facial expressions. It is possible to use the domain-specific knowledge to exploit spatio-temporal sampling using a guided pre-processing technique. Spatio-temporal multi cue networks~\cite{zhou2020spatial} exploit spatial regions of interest by firstly using a branch to estimate the region of interest, then training different networks for each unit. However, applying sampling at the training phase becomes more computationally expensive and requires deeper architectures. Our score-level multi cue fusion approach addresses this problem as described in the next section.

\section{Method} \label{sec3:methodology}


In this section, we describe our method. We firstly describe the mixed convolutional model, follow up with our multi cue sampling process, and finally discuss the score-level fusion method.

\subsection{3D Resnets with Mixed Convolutions}

Mixed convolutional networks are 3D Residual CNNs \cite{tran2015learning}, which use 3D convolutional kernels to process video frames in an end-to-end fashion. Tran et al.~\cite{tran2018resnet3d} investigate the success of 3D CNNs and shares two effective variants with strong empirical results. The first is mixed convolutional networks, and the second is residual bottleneck based 2+1 convolutional networks.  

The mixed CNN variant builds on the plain 2D residual networks, with the difference that the first layers are replaced with 3D convolutional kernels. While the first layers are capable of processing input video directly with 3D convolutional kernels, later layers efficiently model the semantic knowledge using 2D convolutional kernels. Then, a fully connected layer is employed after the final convolutional layer for the video classification task.  

Mixed Convolutional networks are denoted with MC$x$, where $x$ is the number of 3D convolutional layer blocks. Following the baseline, we empirically experiment with different mixed convolutional variants and employ the MC$3$ variant of the mixed convolutional network.

\subsection{Spatial and Temporal Sampling}

The message in a sign gloss is conveyed through manual and non-manual cues. Information is conveyed through the shape and configuration of the hand, body, and face regions. The informative regions and intervals can be sampled with the help of a state-of-the-art pose estimation approach such as OpenPose \cite{cao2019openpose}. Making use of pose estimation allows researchers to filter the entire frame by cropping specific regions according to keypoints, which are hand, face, and upper body keypoints in the case of SLR.

\begin{figure*}[!t]
    \centering
    \includegraphics[width=\textwidth]{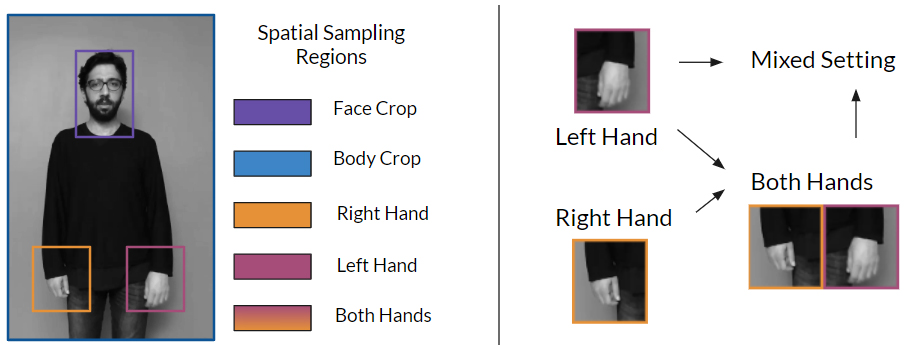}
    \caption{Spatial Sampling operation is visualized. From left to right; cue regions selected for the process, and hand crop settings}
    \label{fig:spatial_fig}
\end{figure*}

\begin{figure*}[!t]
    \centering
    \includegraphics[width=\textwidth]{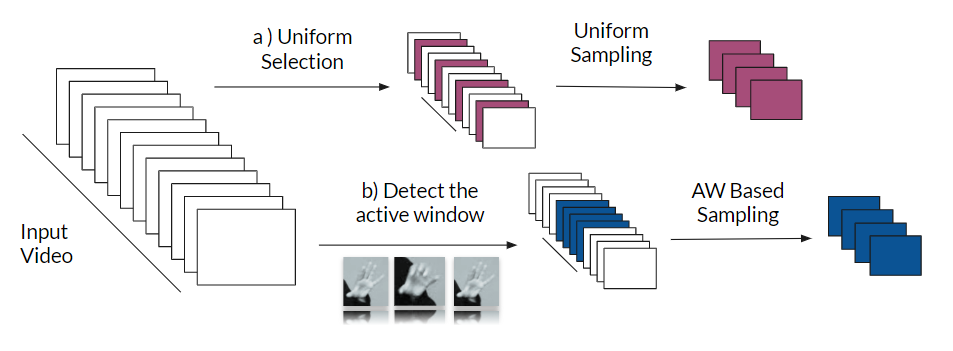}
    \caption{Different temporal sampling operations are shown in the above figure. Selected frames are shown with color. Two branches represent uniform sampling and the Active Window  Based Sampling Process}
    \label{fig:temporal_fig}
\end{figure*}

We would like to sample informative body regions to increase efficiency, and to filter out noise. Our approach is two-fold; (1) We design a SLR system by extracting the body, hand, and face regions by cropping the RGB frame spatially (in Figure \ref{fig:spatial_fig}) using the pose data which was provided in {\"O}zdemir et al.~\cite{ozdemir2020bosphorussign22k}, (2) We focus on the temporal dense regions in which we define the active window as the temporal window where the active hand is moving. Then, we filter out the sparse frames and only feed the network with the frames in the active window (in Figure \ref{fig:temporal_fig}).

Using isolated sign gloss clips guarantees that the temporal sequence is centered on the hand movement. The following steps are used to extract the active window at the center.  
\begin{enumerate}
    \item Use the moving hand detection framework in \ref{spatial-sampling} to detect the active hand(s). 
    \item Define a selected hand as the active hand. If both hands are active, select the dominant hand.
    \item For the selected hand, track hand movements using Euclidean distance. Keep the frame ids of the start of the first-hand movement and end of the last hand-movement.
    \item Define two thresholds $T_S, T_E$. Filter the boundary regions from the start and end frame ids using corresponding thresholds defined earlier, and use extracted frames for the training.
\end{enumerate}

In some videos, the movement is not in the middle of the video. We detect such exception cases by checking the position of the hand relative to the hip. We also filter out segments too short to be a sign.

\subsection{Multi Cue Score Fusion}
Extracting multiple cues from different settings allows each model to build expertise on each cue. Therefore, there is a need to combine the cues of each model by combining weak expert classifiers. Zhou et al. \cite{zhou2020spatial} experiments with distillation at the training time, by training a big scale model consisting of expert components. This has the drawback of increasing model complexity and training time. Simonyan et al. \cite{simonyan2014two} combines different branches while training, but processes the spatial and temporal branch separately at test time using a score fusion approach. They propose firstly direct score fusion via averaging through the network outputs and secondly, training a meta classifier above the extracted features. We follow the former score fusion approach since it has less model complexity and can achieve better run-time performance.

We experiment with two different multi cue fusion settings. First, we apply the averaging operation to the softmax outputs of each cue network results. Secondly, we apply a weighted fusion, where each cue network is weighted by its validation set performance. 

\begin{figure*}[!t]
    \centering
    \includegraphics[width=1\textwidth]{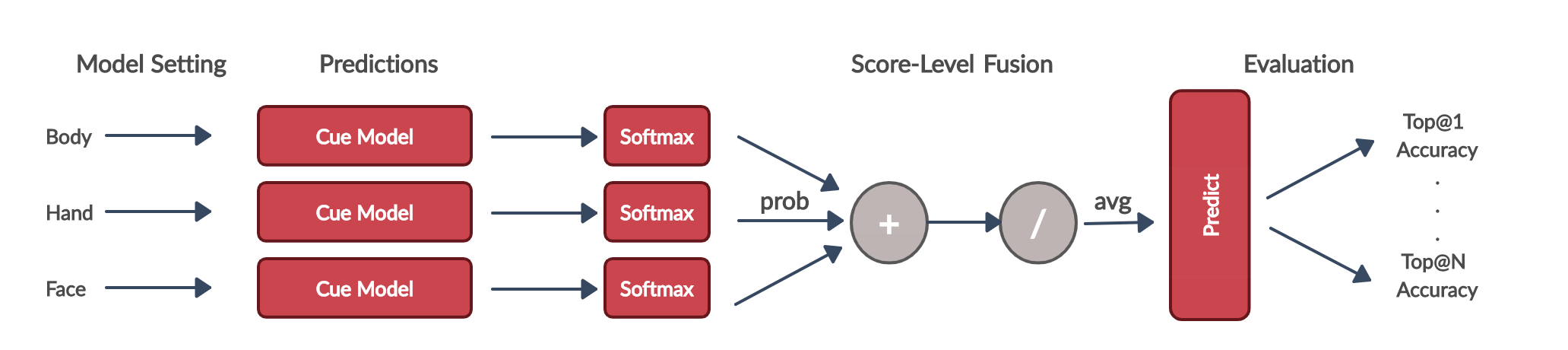}
    \caption{Score-level multi cue fusion operation applied at the test time. Note that cue networks have different test weights even the architecture is same}
    \label{fig:fusion_fig}
\end{figure*}

\section{Experiments} \label{sec4:experiments}

\subsection{Experimental Setup}

\subsubsection{Dataset}

To achieve a competitive experimental setting, and to implement our proposal effectively, we have used a recently published Turkish Isolated SLR dataset BosphorusSign22k ~\cite{ozdemir2020bosphorussign22k}. The dataset contains $6$ different native signers, performing $744$ different sign glosses. Each category is labeled with a sign gloss, that describes the performed sign. The dataset contains over $22,000$ video clips. Authors also share 3D body pose keypoints in Kinectv2 format, and 2D body and hand keypoints obtained from OpenPose \cite{cao2019openpose}.

\subsubsection{Evaluation Metric}
Following the work of {\"O}zdemir et al.~\cite{ozdemir2020bosphorussign22k}, we aim to compete on the sign language classification task. It is described as estimating the corresponding sign gloss for a given input video at test time, and scoring is evaluated in the accuracy of all of the test estimations. Out of all 6 performers, video clips of User 4 is defined as the test set, which is about 1/6 of the total dataset and it includes samples from all of the 744 classes.

\subsubsection{Implementation Details} \label{sec4-3: Implementation Details}
Our experiment setting follows the baseline paper's~\cite{ozdemir2020bosphorussign22k} neural network based experimental setting. We apply the proposed preprocessing pipeline, resize the image into $640\times360$, crop the center square region then resize via bilinear interpolation to achieve $112\times112$ input resolution. Then, we adopt the PyTorch implementation \cite{paszke2017automatic} of the mixed convolutional MC$3$ CNN model which was pretrained on the Kinetics dataset \cite{kay2017kinetics}. In our experiments, we only fine-tuned the last $3$ residual blocks, and apply uniform frame sampling to input video frames. All experiments has been performed with 32 batch size on a Nvidia 1080TI GPU (with 11GB memory).

Our replicated network resulted in $75.23\%$ accuracy, which is more than $3\%$ lower comparing the reported $78.85\%$ accuracy in {\"O}zdemir et al.~\cite{ozdemir2020bosphorussign22k}. We suspect that the difference is caused by randomized states such as optimizer initialization and different hyperparameter choices such as the learning rate.

\subsection{Experimental Results}

\subsubsection{Spatial sampling.} \label{spatial-sampling}

Spatial sampling operation is applied through two phases. First, the cue region is detected, cropped, and optionally concatenated in a multiple cue setting. Secondly, sampling is applied using bilinear interpolation.

\textbf{Body Setting.}
Following the standard SLR pipeline, we crop human body region before training.

\textbf{Hand Setting.}
SLR work suggests that the dominant hand, the most used hand, conveys the most information in communication. To detect the dominant hand in the BosphorusSign22k dataset, we employ a hand motion tracking algorithm. The detection process is achieved by the following:
\begin{enumerate}
\item Detect the Thumb keypoints on each frame,
\item Define the first thumb keypoint on each hand as two anchors,
\item If the following thumb keypoint on the next frames has greater distance than threshold compared to the anchor, conclude the hand as moving.
\end{enumerate}

To compare keypoints for detecting the dominant hand with threshold values (which is predefined as 150 pixels), we use Euclidean distance. Table \ref{tab:hand-settings} provides the detection results on moving hands on BosphorusSign22k dataset. After the detection process, We have seen that signers in the dataset are using their left hands dominantly when performing a sign.

\setlength{\tabcolsep}{4pt}
\begin{table}[!t]
\begin{center}

\caption{Hand spatial sampling settings. First table represents hand activity distribution in the BosphorusSign22k dataset. Second table represents test results of the different hand crop settings and resulting accuracy values}
\label{tab:hand-settings}

\begin{tabular}{lc}
\hline\noalign{\smallskip}
Distribution      & Relative Frequency (\%) \\
\noalign{\smallskip}
\hline
\noalign{\smallskip}
Both Active        & \textbf{66.44}         \\
Only Left Active  & 33.07         \\
Only Right Active & 0.40          \\
\hline
\end{tabular}
\quad
\begin{tabular}{lc}
\hline\noalign{\smallskip}
Crop Setting  &  Accuracy(\%) \\
\noalign{\smallskip}
\hline
\noalign{\smallskip}
Single Hand & 79.13       \\ 
Both Hands  & 85.81       \\ 
Mixed       & \bf{86.25}      \\ 
\hline
\end{tabular}

\end{center}
\end{table}
\setlength{\tabcolsep}{1.4pt}

During signing, only one hand may be active, or both hands may be active. We have adopted three different policies; (1) The single cue setting is applied by selecting the dominant hand in which hand crops with $350\times350$ resolution are obtained around the keypoint \#2. (2) Both cue setting is applied by selecting both hands where hand crops with $175\times350$ resolution are obtained around the Thumb keypoint, and concatenated horizontally. (3) The mixed setting uses the single cue setting when a single hand is active, and uses both cue setting when both hands are active. All three settings are followed by downsampling with bilinear interpolation. Experimental results are provided at the right-hand side of the Table \ref{tab:hand-settings}.

\textbf{Face Setting.}
Signers often have cues with facial expressions or lip movements (mouthings) that can give hints about the sign gloss. For this purpose, we have also experimented on a face setting where we crop the entire face from frames. To crop the face, we have used the Nose keypoints which are provided with the Openpose \cite{cao2019openpose} keypoints. After cropping the face, we resize them to $200\times200$ resolution. 

\noindent\textbf{Score-Level Fusion}
We follow the insight that the different cue models can capture a different subset of features, which can lead to better results when combined effectively. Standard fusion is applied by averaging softmax outputs as in ~\cite{simonyan2014two}. In the weighted setting, we have applied weights to each model proportional to their validation accuracy via standard multiplication. Table \ref{tab:spatiotemporal-result} provides the result of the fusion.

\setlength{\tabcolsep}{4pt}
\begin{table}[!t]
\begin{center}
\caption{Classification accuracy results of the sampling and fusion settings. Three different settings are provided in the table. From left to right, (1) Single cue spatial sampling results, (2) Active Window Based Temporal Sampling applied to each crop, and (3) Spatial\&Temporal settings are combined in one setting. Note that the bottom two rows include the fusion result of the above three models in each setting.}
\label{tab:spatiotemporal-result}
\begin{tabular}{lcc|cc|cc}
\hline\noalign{\smallskip}
 & \multicolumn{2}{c}{Spatial}  & \multicolumn{2}{|c}{Temporal} & \multicolumn{2}{|c}{S\&T Combined}\\
\noalign{\smallskip}
\hline
\noalign{\smallskip}
Setting & Acc@1   & Acc@5 & Acc@1   & Acc@5 & Acc@1   & Acc@5  \\
\noalign{\smallskip}
\hline
\noalign{\smallskip}
Body & 75.73 & 93.88 & 81.83 & 96.02 & 86.91 & 98.17 \\
Hand & 86.25 & 97.61 & 88.70 & 97.59 & 91.73 & 98.72 \\
Face & 24.27 & 44.45 & 37.00 & 57.89 & 39.12 & 59.33 \\
Fusion & 90.63 & 98.92 & 93.88 & \textbf{99.65} & 94.47& \textbf{99.78} \\
Weighted Fusion & \textbf{92.18} & \textbf{99.27} & \textbf{94.03} & 99.56 & \textbf{94.94} & 99.76 \\
\hline
\end{tabular}
\end{center}
\end{table}
\setlength{\tabcolsep}{1.4pt}

\subsubsection{Temporal Sampling}
Standard SLR training pipeline involves using the standard uniform frame sampling. We propose the active window based temporal sampling, applied by firstly extracting the dense cue regions before applying the uniform selection. Active window is detected as the part that the active hand is moving and discard the rest of the temporal information.

We used double thresholding for finding the active window. We have found that the start threshold $T_S=90$, and the end threshold $T_E=50$ generates competitive empirical results. Using the temporal sampling framework, we have successfully segmented the active window for each video. Then, we applied uniform sampling along with our standard training pipeline. Experimental results can be seen in Table \ref{tab:spatiotemporal-result}.


\subsubsection{Spatio-Temporal Sampling}
We applied active window based temporal sampling on top of the spatial multi cue regions. Our experiments have shown that the final spatio-temporal sampling framework has improved on both single cue settings. With the addition of score-level fusion, test accuracy reached to $94.94\%$, which is the best result in all proposed settings as seen in the Table \ref{tab:spatiotemporal-result}.

Our best setting provides $16.09\%$ improvement on our baseline neural network setting \cite{ozdemir2020bosphorussign22k}. We also managed to improve their previous best hand-crafted result with $6.41\%$ accuracy rate. Whereas the previous best method uses more than ten times bigger input spatial resolution ($640\times360$), complicated hand-crafted methods~\cite{wang2013action} and a second stage SVM classifier, our approach only contains a 3D CNN and a sampling pipeline. Comparison with the baseline results is shown in Table \ref{tab:baseline}.

\setlength{\tabcolsep}{4pt}
\begin{table}[!t]
\centering
\caption{Comparison with the baseline approaches IDT and MC3\_18 model.}
\label{tab:baseline}
\begin{tabular}{lcc}
\noalign{\smallskip}
\hline
\noalign{\smallskip}
Method & Acc@1 & Acc@5 \\ 
\noalign{\smallskip}
\hline
\noalign{\smallskip}
Baseline IDT \cite{ozdemir2020bosphorussign22k} & 88.53 & - \\
Baseline MC3\_18 \cite{ozdemir2020bosphorussign22k} & 78.85 & 94.76 \\
Weighted Fusion - S\&T Combined & \textbf{94.94} & \textbf{99.76} \\
\noalign{\smallskip}
\hline
\noalign{\smallskip}
\end{tabular}%
\end{table}
\setlength{\tabcolsep}{1.4pt}

\section{Discussion and Analysis} \label{sec5:discussion}

\begin{figure*}[!t]
    \centering
    \includegraphics[width=\textwidth]{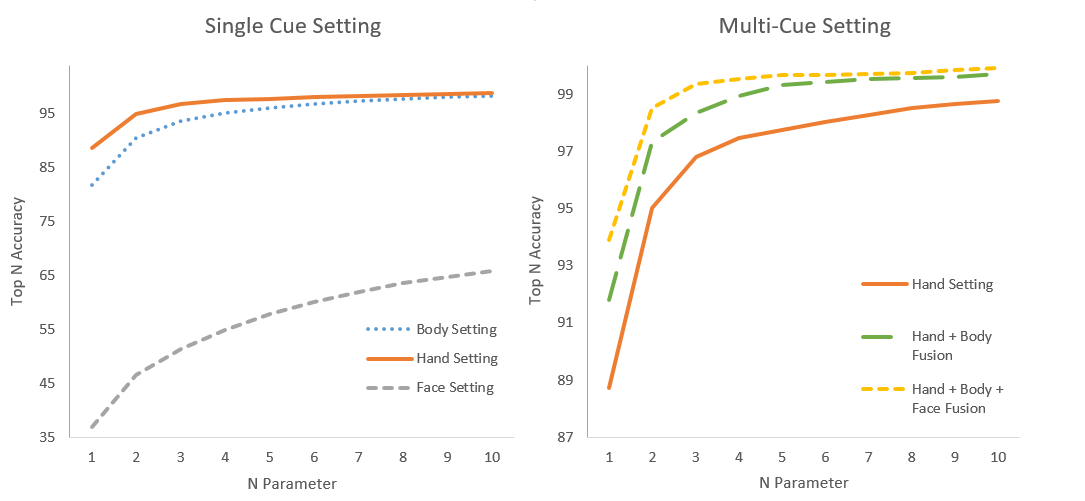}
    \caption{Comparison of the models in the Top-N accuracy setting. The horizontal axis denotes the increasing N value, and the vertical axis denotes the accuracy value. First plot shows the single cue setting comparison, and second plot shows the multi cue setting additive comparison. Despite the difference in single setting performance, each cue boosts the fusion results.}
    \label{fig:topn-combined}

\end{figure*}

Accuracy lacks informativeness when considering whether the fusion will be beneficial or not. Top-N Accuracy measures how often the Top-N ranks contain the correct class. In our experiments, we also analyze Top-5 Accuracy along with Top-1 Accuracy. Top-N Accuracy results will increase with an increasing N, and are expected to be settled to 1 when N approaches to the maximum class number. Our Top-N accuracy analysis can be seen in Figure \ref{fig:topn-combined}.

In the plot on the left-hand side, we report Top-N accuracy of the individual cues. Our analysis shows that hand cue yields the best performance, which is followed by the body cue. In both, there is a sharp increase between ranks 1 and 2. This shows that in a large number of cases, although the correct class fails to be predicted, it is the runner-up. This explains why the fusion is beneficial. Although the Top-N accuracy of the face cue is much lower, it is still beneficial for fusion. 

Top-N accuracy of the muti-cue fusion is given in the right-hand side of Figure \ref{fig:topn-combined}. We start by the hand model, then include the body model, and finally add the face model to the mix. This analysis allows us to see the cumulative progress over different fusion models. We observe that the Top-2 accuracy of hand alone is higher than Top-1 accuracy of both fusion settings. We believe that this observation is why the weighted fusion outperforms score fusion, and shows that more advanced models can attain higher performance.

\subsection{Spatial Ablation Study}

To analyse which cue benefits the fusion results the most, we have performed score fusion to all combination pairs of cue settings. According to this ablation study, we were able to observe the effect of each cue to the overall fusion. For example, to find the effect of the face model, we subtract the Body+Hand setting from the Body+Hand+Face setting. Table \ref{tab:fusion_ablation} shows the results of the ablation study. In our analysis, we can see that hand cue has the most effect on the fusion by $9.22\%$ which is followed by the body model with $5.18\%$.


\setlength{\tabcolsep}{4pt}
\begin{table}[!t]
\begin{center}
\caption{Effects of excluding individual cue units from the final fusion model. Using the different two cue settings and their performance, we infer to the excluded setting and its effect on the final mix.}
\label{tab:fusion_ablation}
\begin{tabular}{lclc}
\hline\noalign{\smallskip}
Setting & Accuracy & Excluded Cue & Effect (\%)\\
\noalign{\smallskip}
\hline
\noalign{\smallskip}
Body + Hand  & 91.80 & Face  & 2.08 \\
Body + Face   & \textbf{84.66} & Hand & \textbf{9.22} \\
Hands + Face  & 88.70 & Body  & 5.18 \\
\hline
\end{tabular}
\end{center}
\end{table}
\setlength{\tabcolsep}{1.4pt}

We have provided an analysis of the two most effective cues by comparing the gloss based performance. As a comparison metric, we adopted the F1-score, which should be more representative of false positives and false negatives, thus is more suitable for the gloss based evaluation.

\textbf{Gloss Based Cue Comparison} We share the top ten sign glosses that the hand cue model has a major advantage compared to the body cue model in Table \ref{tab:hand_advantage}. 

\setlength{\tabcolsep}{4pt}
\begin{table}[!t]
\begin{center}
\caption{F1-score comparison for the top ten sign glosses that hand sampling outperforms body sampling. (Sorted in the alphabetical order)}
\label{tab:hand_advantage}
\begin{tabular}{lccc||lccc}
\hline\noalign{\smallskip}
Sign Gloss & Hand  & Body  & Fusion & Sign Gloss & Hand  & Body  & Fusion\\
\noalign{\smallskip}
\hline
\noalign{\smallskip}
Aspirin & 0.62 & 0.00 & 0.67       & 
Internet\_2 & 1.00 & 0.33 & 1.00      \\
Deposit(v)\_2 & 0.89 & 0.25 & 1.00 &
Noon & 0.91 & 0.33 & 1.00          \\
Exchange(v) & 0.57 & 0.00 & 1.00      &
Shout(v)\_2 & 0.91 & 0.33 & 0.91\\
Head & 0.89 & 0.29 & 1.00          &
Sleep(v) & 0.62 & 0.00 & 0.67      \\
Identify(v) & 0.89 & 0.00 & 1.00      &
Turn(v) & 1.00 & 0.40 & 0.89        \\
\hline
\end{tabular}
\end{center}
\end{table}
\setlength{\tabcolsep}{1.4pt}

\begin{figure*}[!t]
    \centering
    \includegraphics[width=\textwidth]{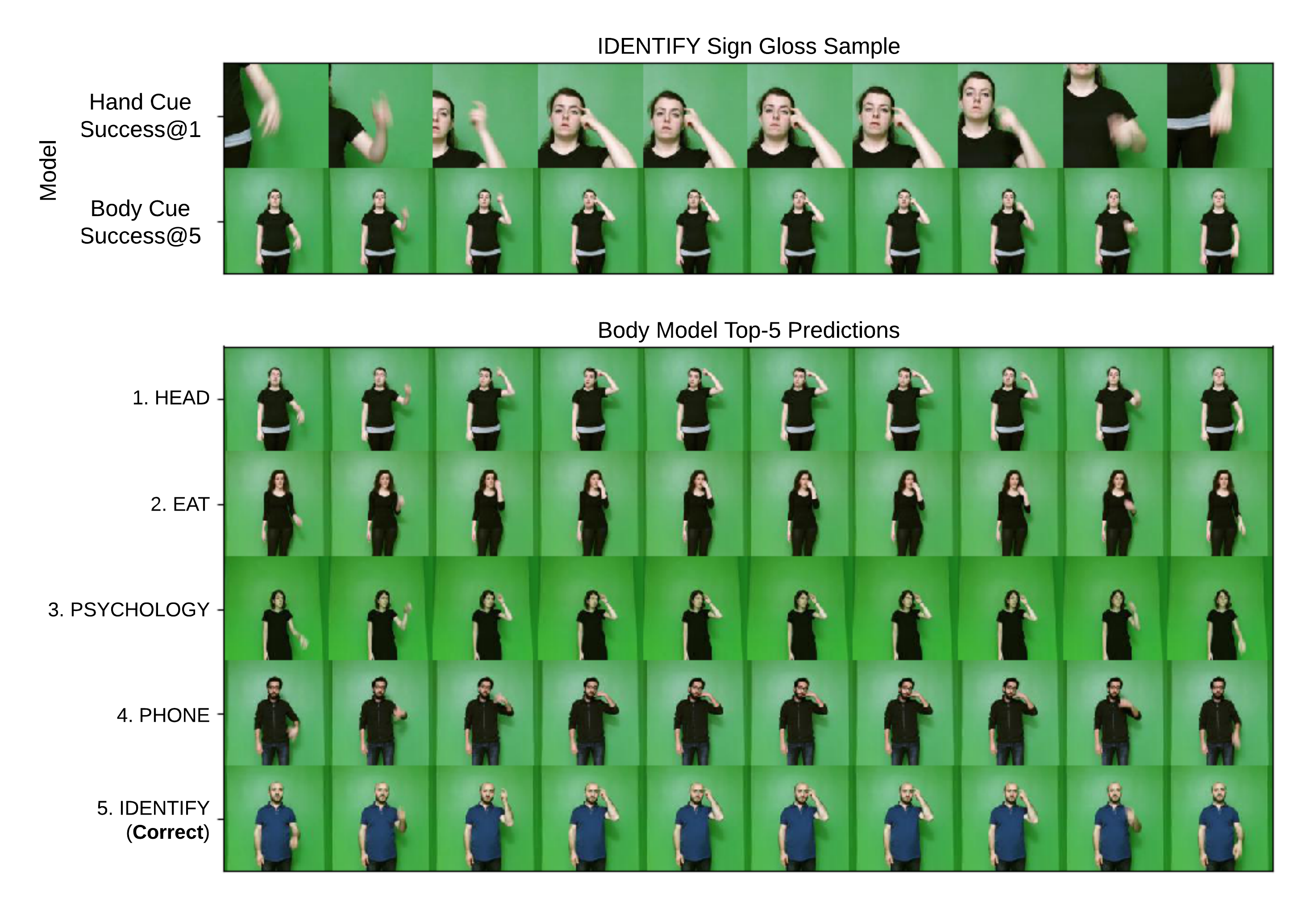}
    \caption{Class confusions of IDENTIFY(v) sign gloss for the body cue model}
    \label{fig:hand_over_body}
\end{figure*}

In Figure \ref{fig:hand_over_body}, we provide detailed analysis for the IDENTIFY(v) sign gloss. IDENTIFY(v) sign gloss is performed by using only the left hand, touching the head with the index finger, and the rest of the fingers are on the semi-open position. In this particular example, 
the body cue model only achieves success in the 5th guess, while the hand cue model has the correct prediction. Additionally, Figure \ref{fig:hand_over_body} also shows that misclassifications of the body cue model which are HEAD, EAT, PSYCHOLOGY, and PHONE sign glosses. We inspect each confusion as follows:


\begin{itemize}
    \item \textbf{HEAD} sign differs from IDENTIFY(v) with the close position on all fingers other than the index finger. 
    \item \textbf{EAT} sign is performed by moving the left hand close to the mouth and with all fingers are in a closed position.  
    \item \textbf{PSYCHOLOGY} and \textbf{PHONE} sign glosses are performed with the left hand that and have open and semi-closed hand shapes, respectively.
\end{itemize}

By evaluating the confused cases, we have concluded that the hand model has an advantage of capturing hand shape information which was possibly due to increased spatial resolution of the hand region.

\textbf{Effect of the Score-Level Fusion.} 
We share our fusion result of the body and hand cue models in Table \ref{tab:hand_advantage}. Data has shown that the fusion model successfully captures the hand cue features. We have also seen that the fusion model even outperforms both single cue models in $7$ out of $10$ glosses. 


\subsection{Analysis of Method on Types of Gestures Recognized}
To further analyze the types of signs which the proposed method performs well and fails, we have labeled the $744$ sign glosses in the dataset according to specific sign attributes. The sign classes are grouped into categories such as one-handed signs, two-handed signs, mono-morphemic signs, compound signs, and signs involving repetitive and circular movements of the hands. 

Table \ref{tab:analysis} summarizes the analysis: The experiments are performed using temporal sampling with the best performing mixed convolution approach. Attribute-wise accuracy scores are calculated using the test set samples belonging to the classes containing the selected attributes. Overall, the accuracy scores in Table \ref{tab:analysis} demonstrate that for nearly all the subsets in the dataset, hand, body, and face-based features show consistency in their relative performance. 

\setlength{\tabcolsep}{4pt}
\begin{table}[!t]
\begin{center}
\caption{Analysis of the temporal sampling based recognition approach with respect to signs with certain grammatical sign attributes: one-handed signs, two-handed signs, mono-morphemic signs, compound signs, and signs involving repetitive and circular movements, respectively}
\label{tab:analysis}
\begin{tabular}{p{1.5cm}p{.9cm}p{.9cm}p{.9cm}p{.9cm}p{.9cm}p{.9cm}p{.9cm}p{.9cm}p{.9cm}}
      & \multicolumn{9}{c}{Number of Classes with Selected Attribute}  
      \\
      \hline
      & 234        & 510        & 75       & 669          & 457        & 287            & 375     & 369      & 744     \\
      \hline
      & One Handed & Two Handed & Circ. & Not Circ. & Rep. & Not Rep.& Mono & Comp. & All     \\
\hline
Body  & 72.94    & 86.10    & 86.43  & 81.33      & 80.77    & 83.55        & 77.40 & 86.48  & 81.83 \\
Hand  & 83.78    & 91.07    & 91.40  & 88.41      & 87.54    & 90.59        & 85.36 & 92.24  & 88.70 \\
Face  & 45.33    & 33.01    & 29.41  & 37.82      & 36.39    & 37.99        & 33.79 & 40.52  & 37.00 \\
Fusion & \textbf{91.82}    & 94.86    & \textbf{96.15}  & 93.63      & \textbf{93.45}    & 94.57        & 92.08 & 95.78  & 93.88 \\
W.Fusion & 90.87    & \textbf{95.55}    & 95.70  & \textbf{93.85}      & 93.37    & \textbf{95.09}        & \textbf{92.21} & \textbf{95.96}  & \textbf{94.03} \\
\hline
\end{tabular}
\end{center}
\end{table}
\setlength{\tabcolsep}{1.4pt}

Looking at the results for different attributes one by one, we can see that signs involving two moving hands are better recognized than the one-handed sign glosses in the dataset. The performance difference can be explained by the fact that in one-handed signs, the weight of handshape may be more critical than the two-handed signs. The relative positioning and appearance of both hands, which is more apparent, may be easier to represent for the neural network. 

Secondly, compound signs have a greater recognition accuracy than mono-morphemic signs ($95.96\%$ vs $92.21\%$). Considering the number of signing hands, the amount of additional information in the form of consecutive morphemes present in an isolated sign makes recognition easier, thus improving the performance system. From this result, we can infer that the method's representation power is higher when a sign is greater in length and contains different hand shape and position combinations.

Looking at repetitive gestures, we see a $1.6\%$ improvement in accuracy when the signs do not contain repetitive hand gestures. The issue with repetitions, which we can attribute to this difference, is that the temporal and spatial forms of repetitions are more prone to differ between performances and users, in comparison to the static hand shape parts of the signs that follow specific rules. 

Finally, we take a look at circular signs, which include circular hand and arm movements, which involve at least one entire rotation. These signs are dynamic signs where the hands do not stop while presenting a handshape. As these signs do not conform to the movement-hold phonological model of sign languages \cite{liddell1989american}, representing them by choosing temporal frames is more complicated, reducing the effectiveness of keyframe based approaches \cite{kindiroglu2019taf}. Overall, the method performs well with circular signs, making fusion attempts with methods focusing more on the handshape of signs promising future leads.   

\section{Conclusion} \label{sec6:conclusion}

In this paper, we have proposed a score-level multi cue fusion approach for the Isolated SLR task. Unlike the previous work~\cite{ozdemir2020bosphorussign22k,kindiroglu2019taf}, we focused on both spatial and temporal cues. We employed 3D Residual CNNs \cite{tran2018resnet3d}, and trained different models as an expert on the single cue. We distilled the expert knowledge using the weighted and unweighted score-Level fusion. In our experiments, we have seen that our approach has outperformed the baseline results on the BosphorusSign22k Turkish Isolated SL dataset~\cite{ozdemir2020bosphorussign22k}.

We have provided the single cue and multi cue Top-N accuracies to demonstrate incremental performance gain with each cue. Our gloss-level study shows that each cue model has specific expertise and provides an indispensable knowledge source to the fusion model. Our analysis of sign gloss attributes hints that the method performs better on temporally more complex signs with two-handed gestures, while performing comparatively worse on mono-morphemic gestures with a single hand. For that reason, the primary approach to improving performance lies in improving hand shape recognition. Possible strategies involve increasing model depth, finding better optimization techniques, or increasing the model input size. We hope that this work will extend the SLR cues into other Sign Language problems, help progress in unresolved SL tasks such as translation, and help uncover language-independent cues.
Prob

\section*{Acknowledgement}
This work has been supported by the TUBITAK Project No. 117E059 and TAM Project No. 2007K120610 under the Turkish Ministry of Development.

%
%
\clearpage
\bibliographystyle{splncs04}
\bibliography{egbib}
\end{document}